\documentclass{bmvc2k}
\usepackage{color}
\usepackage{algorithm}
\usepackage{algpseudocode}
\usepackage{amsmath,amssymb,amsfonts}
\usepackage{booktabs}       %
\usepackage{multirow, varwidth}
\usepackage{mathtools}
\usepackage{epsfig}
\usepackage{verbatim}
\usepackage{xr}
\usepackage{flushend}
\usepackage{subcaption}

\usepackage{graphicx}
\usepackage{url}
\usepackage{enumitem}
\setlist{nosep} %
\usepackage[table]{xcolor}         %
\usepackage{kotex}
\usepackage{colortbl}

\DeclareRobustCommand\onedot{\futurelet\@let@token\@onedot}
\def\onedot{.} %
\def\eg{\emph{e.g}\onedot, } 
\def\ie{\emph{i.e}\onedot, } 
\def\cf{\emph{c.f}\onedot, }

\def\etal{\emph{et al}\onedot}

\newcommand{\mytilde}{\raise.17ex\hbox{$\scriptstyle\mathtt{\sim}$}}

\title{
Beyond Softmax: Dual-Branch Sigmoid Architecture for Accurate Class Activation Maps
}
\addauthor{Yoojin Oh}{mydianaoh@ewha.ac.kr}{1}
\addauthor{Junhyug Noh}{junhyug@ewha.ac.kr}{1}
\addinstitution{
 Department of Artificial Intelligence\\
 Ewha Womans University\\
 Seoul, Republic of Korea
}
\runninghead{Oh \& Noh}{Beyond Softmax}
\def\eg{\emph{e.g}\bmvaOneDot}

\def\etal{\emph{et al}\bmvaOneDot}
\begin{document}

\maketitle

\begin{abstract}
Class Activation Mapping (CAM) and its extensions have become indispensable tools for visualizing the evidence behind deep network predictions. However, by relying on a final softmax classifier, these methods suffer from two fundamental distortions: \emph{additive logit shifts} that arbitrarily bias importance scores, and \emph{sign collapse} that conflates excitatory and inhibitory features. We propose a simple, architecture-agnostic \emph{dual-branch sigmoid} head that decouples localization from classification. Given any pretrained model, we clone its classification head into a parallel branch ending in per-class sigmoid outputs, freeze the original softmax head, and fine-tune only the sigmoid branch with class-balanced binary supervision. At inference, softmax retains recognition accuracy, while class evidence maps are generated from the sigmoid branch -- preserving both magnitude and sign of feature contributions. Our method integrates seamlessly with most CAM variants and incurs negligible overhead. Extensive evaluations on fine-grained tasks (CUB-200-2011, Stanford Cars) and WSOL benchmarks (ImageNet-1K, OpenImages-30K) show improved explanation fidelity and consistent Top-1 Localization gains -- without any drop in classification accuracy. Code is available at \url{https://github.com/finallyupper/beyond-softmax}.
\end{abstract}

\section{Introduction}
\label{sec:intro}

Deep neural networks have achieved remarkable success across a wide range of visual recognition tasks, from fine-grained classification~\cite{wah2011cub} to large-scale object detection~\cite{Imagenet:Russakovsky2015}.
However, their decision processes remain largely opaque, limiting trust in safety-critical domains such as medical imaging and autonomous driving.
Class Activation Mapping (CAM)~\cite{zhou2016learning} was proposed to bridge this gap by projecting the weighted sum of high‐level feature maps back onto the input image, yielding a heatmap that highlights discriminative regions for a given class.
While effective, CAM requires architectural constraints (Global Average Pooling \& fully-connected layers) and exposes only linear combinations of feature maps.

Grad-CAM~\cite{selvaraju2017grad} and Grad-CAM++~\cite{chattopadhay2018grad} generalize CAM to arbitrary network architectures -- be they convolutional backbones with deep MLP heads or transformer encoders with \texttt{[CLS]} tokens -- by using gradients of the target logit to compute per‐channel importance scores.
A host of further variants have since been developed: Score-CAM~\cite{wang2020score} and Ablation-CAM~\cite{ramaswamy2020ablation} eschew gradients entirely, measuring score perturbations under masking or ablation; Layer-CAM~\cite{jiang2021layercam} and Eigen-CAM~\cite{muhammad2020eigen} fuse multi-layer activations or principal components; and attention-based methods directly visualize ViT attention maps~\cite{Rollout:Abnar2020,TSCAM:Gao2021}.

Despite these advances, most CAM‐style explanations inherit two fundamental distortions from the ubiquitous softmax classifier: (i) \emph{additive logit shift}, whereby adding a constant to all class logits leaves softmax probabilities unchanged but arbitrarily biases importance scores; and (ii) \emph{sign collapse}, whereby softmax depends only on relative ordering of logits and discards the absolute sign of feature contributions.  We show in Section~\ref{sec:softmax_distortions} that these artifacts can dramatically mislead localization maps.

To overcome these issues, we decouple localization from classification via a lightweight, architecture-agnostic \emph{dual-branch sigmoid} head. 
Starting from any pretrained model, we clone its classification head into a parallel sigmoid branch -- freeze the original softmax branch and fine-tune only the sigmoid branch with class-balanced binary supervision. 
At inference, we retain softmax for recognition and use the sigmoid branch to generate class evidence maps with absolute, signed feature importance. 
This plug-in approach works across any CAM variants and delivers consistent gains in two scenarios: (1) \emph{explanation fidelity} on fine-grained tasks (CUB-200-2011~\cite{wah2011cub} and Stanford Cars~\cite{krause2013cars}), reducing Average Drop and boosting \% Increase in Confidence and (2) \emph{weakly supervised object localization} on ImageNet-1K~\cite{Imagenet:Russakovsky2015} and OpenImages-30K~\cite{OpenImages:Benenson2019} -- improving Top-1 Loc, MaxBoxAccV2, and PxAP without harming classification -- across multiple CAM methods.

In summary, our contributions are:
\begin{itemize}
  \item A formal analysis of two softmax-induced distortions -- \emph{additive logit shift} and \emph{sign collapse} -- that undermine CAM-style explanations.
  \item A dual-branch sigmoid head that restores absolute, signed feature importance for arbitrary classifier architectures.
  \item Extensive experiments showing enhanced explanation fidelity on CUB-200-2011 and Stanford Cars, and WSOL improvements on ImageNet-1K and OpenImages-30K, all with negligible overhead.
\end{itemize}

\section{Related Work}
\label{sec:related_work}

\noindent\textbf{Gradient-Based Methods.}
Early pixel‐level saliency maps compute the gradient of a class score with respect to each input pixel~\cite{simonyan2013deep}.  
SmoothGrad~\cite{smilkov2017smoothgrad} reduces noise by averaging over noisy inputs, and Integrated Gradients~\cite{sundararajan2017axiomatic} accumulates gradients along a path from a baseline.  
CAM~\cite{zhou2016learning} first extended saliency to feature maps via Global Average Pooling and a linear head.  
Grad‐CAM~\cite{selvaraju2017grad} and Grad‐CAM++~\cite{chattopadhay2018grad} generalized this to arbitrary architectures by averaging channel gradients, with Grad‐CAM++ using higher‐order derivatives for multiple-instance localization.  
Layer‐CAM~\cite{jiang2021layercam} fuses activation-gradient products across layers, and XGrad-CAM~\cite{fu2020axiom} derives weights from axiomatic properties.

\smallskip
\noindent\textbf{Gradient-Free and Perturbation-Based Methods.}
To avoid gradient saturation, Score-CAM~\cite{wang2020score} uses the forward pass with feature‐map masks to measure class scores, and Ablation-CAM~\cite{ramaswamy2020ablation} quantifies score drops when ablating channels.  
RISE~\cite{petsiuk2018rise} averages random input masks, while Extremal Perturbations~\cite{fong2019understanding} optimize sparse masks to highlight important regions.  
These perturbation methods often yield sharper maps but require multiple forward passes.

\smallskip
\noindent\textbf{Statistical and Structural Methods.}
Eigen-CAM~\cite{muhammad2020eigen} leverages the first principal component of the feature tensor as a heatmap.  
Transformer‐specific approaches -- Attention Rollout~\cite{Rollout:Abnar2020} and TS-CAM~\cite{TSCAM:Gao2021} -- visualize aggregated self‐attention.  
Relevance propagation techniques such as LRP~\cite{bach2015pixel} and DeepLIFT~\cite{shrikumar2017learning} distribute the prediction score back through network layers.

\smallskip
\noindent\textbf{Weakly‐Supervised Object Localization.}
WSOL methods seek to predict object bounding boxes or masks using only image‐level labels.
Early works~\cite{HaS:Singh2017, Wei2017, ACoL:Zhang2018, ADL:Choe2019} train models by adapting CAM heatmaps to cover entire objects.  
More recent work, such as Rethinking CAM~\cite{bae2020rethinking}, improves the CAM pipeline itself through refined thresholds and evaluation protocols.  
Unlike post‐hoc explanations, WSOL integrates localization objectives or auxiliary branches during training to directly encourage complete object coverage.

\smallskip
Most CAM‐style methods derive channel or pixel weights from softmax logits (or their perturbations) and thus inherit \emph{additive logit shifts} and \emph{sign collapse} distortions. 
In contrast, our dual‐branch sigmoid head trains per‐class binary classifiers to recover absolute magnitude and polarity of feature contributions, and plugs seamlessly into any CAM or WSOL pipeline to produce more faithful localization maps.

\section{Our Approach}
\label{sec:approach}

We propose a dual-branch sigmoid head that decouples localization from classification by restoring absolute, signed feature importance in CAM-style heatmaps.
Our approach unfolds in three parts: 
first, Sec.~\ref{sec:prelim} reviews CAM fundamentals; 
next, Sec.~\ref{sec:softmax_distortions} analyzes softmax-induced distortions; 
and finally, Sec.~\ref{sec:method} details our dual-branch sigmoid architecture along with its training and inference procedures.

\subsection{Preliminaries: CAM Definitions}
\label{sec:prelim}
Given an input image \(\mathbf{x}\), let $F_i\in\mathbb{R}^{P\times Q}$ be the $i$-th channel feature map of the final convolutional layer, with $i=1,\dots,N$. A classification head $h$ maps the feature tensor $\mathbf{F}\in\mathbb{R}^{N\times P\times Q}$ to logits
\begin{equation}
  \ell_k = h_k(\mathbf{F}), \quad k=1,\dots,C,
  \label{eq:logits}
\end{equation}
with softmax probabilities:
\begin{equation}
  y_k = \frac{\exp(\ell_k)}{\sum_{j=1}^C \exp(\ell_j)}.
  \label{eq:softmax}
\end{equation}

\paragraph{Vanilla CAM.}
In the original CAM~\cite{zhou2016learning}, $h$ is a Global Average Pooling (GAP) layer followed by a fully-connected (FC) layer.

Specifically,
\begin{equation}
  \ell_k = \sum_{i=1}^N w_{i,k}\,\bar F_i + b_k,
  \qquad
  \bar F_i \triangleq \tfrac{1}{PQ}\sum_{p=1}^P\sum_{q=1}^Q F_i^{(p,q)}.
  \label{eq:cam_logit}
\end{equation}
where $w_{i,k}$ and $b_k$ are the FC weights and bias.
The (linear) class activation map for class $k$ is
\begin{equation}
  M_k(p,q) \;=\; \sum_{i=1}^N w_{i,k}\,F_i^{(p,q)}.
  \label{eq:cam}
\end{equation}

\paragraph{Gradient-Based CAM.}
To remove architectural constraints, gradient-based CAM variants~\cite{selvaraju2017grad, chattopadhay2018grad} first compute intermediate importance scores:
\begin{equation}
  \alpha_{i,k}
  = \frac{1}{PQ}\sum_{p=1}^P\sum_{q=1}^Q
    \frac{\partial \ell_k}{\partial F_i^{(p,q)}},
  \label{eq:gradcam_importance}
\end{equation}
which are then post‐processed -- \eg, directly as \(w_{i,k}=\alpha_{i,k}\) in Grad-CAM, refined by higher‐order derivatives in Grad-CAM++~\cite{chattopadhay2018grad}, or fused across layers in Layer-CAM~\cite{jiang2021layercam} -- to obtain the final weights for Eq.~\eqref{eq:cam}.

Many variants additionally apply an elementwise ReLU to obtain a non-negative heatmap:
\begin{equation}
  M_k(p,q) \;\leftarrow\; \mathrm{ReLU}\!\big(M_k(p,q)\big).
  \label{eq:relu_convention}
\end{equation}
We adopt this ReLU-applied map as the default for visualization and evaluation; however, for analytical clarity (e.g., in Sec.~\ref{sec:softmax_distortions}) we refer to the \emph{linear} score $M_k$ in Eq.~\eqref{eq:cam} without the ReLU.

\paragraph{Other Variants.}
Beyond vanilla CAM and gradient-based extensions, many CAM methods~\cite{wang2020score, ramaswamy2020ablation, muhammad2020eigen, fu2020axiom, zhang2025finer} adhere to the same two‐step recipe. 
First, they derive per‐channel importance scores \(w_{i,k}\) from the model's softmax logits or scores -- using gradients, higher‐order derivatives, or perturbations of inputs or feature maps. 
Second, they linearly combine these weights with the feature maps to yield the final heatmap as Eq.~\eqref{eq:cam}.
For example, Score‐CAM~\cite{wang2020score} masks the input with each feature map in turn and measures the resulting change in logit
\begin{align}
  \alpha_{i,k}
  &= \ell_k(\mathbf{x}\odot \mathrm{mask}_i) \;-\; \ell_k(\mathbf{x}),
  \label{eq:score_importance}
\end{align}  
then normalizes \(\{\alpha_{i,k}\}\) to obtain \(w_{i,k}\).

\subsection{Softmax-Induced Distortions}
\label{sec:softmax_distortions}

As reviewed above, all CAM variants ultimately form a heatmap by linearly combining feature maps with per-channel weights (Eq.~\eqref{eq:cam}). 
For this linear combination to be semantically valid, the weights should satisfy two tacit requirements: 
(i) \emph{relative magnitude semantics} -- $w_{i,k}$ should reflect how important channel $i$ is for class $k$ so that weights are comparable across channels; and 
(ii) \emph{sign semantics} -- the sign of $w_{i,k}$ should indicate whether the corresponding activation contributes evidence (positive) or inhibition (negative) to class $k$.

\begin{figure*}[t!]
\begin{center}
\includegraphics[width=0.9\textwidth]{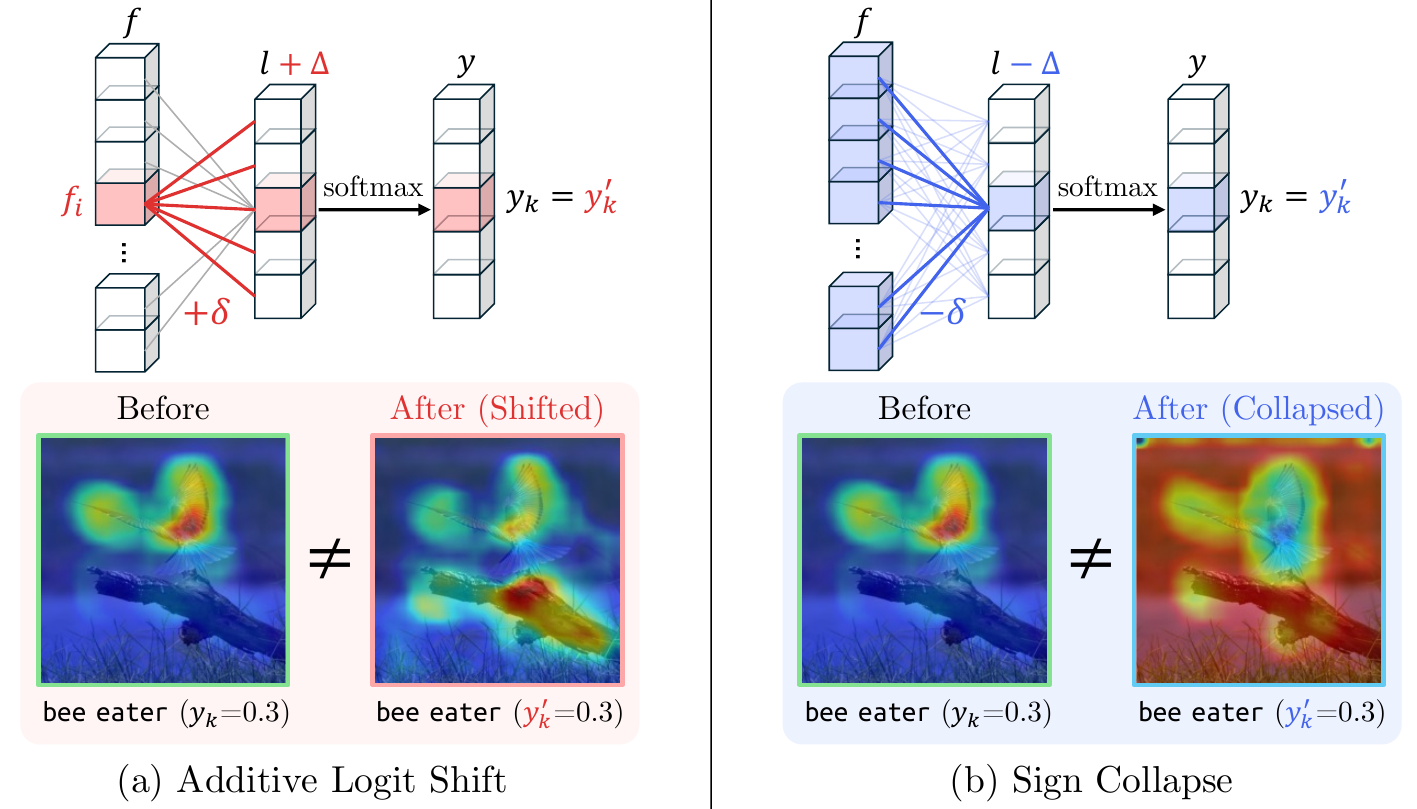}
\end{center}
\caption{
Softmax-induced distortions in CAM-based localization. 
(a) \emph{Additive Logit Shift}: adding a constant $\delta$ to all feature weights leaves the softmax probability $y_k$ unchanged but disproportionately amplifies feature $f_i$ in the heatmap. 
(b) \emph{Sign Collapse}: subtracting $\delta$ flips formerly positive feature weights to negative without affecting $y_k$, causing previously highlighted regions to vanish. 
In both cases, identical classification outputs produce drastically different localization maps.
}
\label{fig:problem}
\end{figure*}

However, when weights are derived from \emph{softmax}-based scores, softmax's invariances break these assumptions: \emph{additive logit shifts} arbitrarily bias magnitudes, and \emph{sign collapse} conflates evidence and inhibition.
Figure~\ref{fig:problem} illustrates these effects with concrete examples.

\paragraph{1) Additive Logit Shift.}
Softmax is invariant to adding the same constant to all class logits.
Let $\boldsymbol{\ell}\in\mathbb{R}^C$ and $\mathbf{1}\in\mathbb{R}^C$ be the all-ones vector.
For any $\Delta\in\mathbb{R}$, define $\boldsymbol{\ell}'=\boldsymbol{\ell}+\Delta\,\mathbf{1}$. Then
\begin{equation}
  y'_k \;=\; \frac{e^{\ell_k+\Delta}}{\sum_{c=1}^C e^{\ell_c+\Delta}}
          \;=\; \frac{e^{\ell_k}e^{\Delta}}{e^{\Delta}\sum_{c=1}^C e^{\ell_c}}
          \;=\; y_k.
  \label{eq:shift_invariance}
\end{equation}

Now apply a \emph{uniform} shift to all FC weights connected to the $i$-th feature map across classes:
\[
  w'_{i,k}=w_{i,k}+\delta\quad(\forall k),\qquad
  w'_{j,k}=w_{j,k}\ \ (j\neq i).
\]
Under the GAP\(+\)FC head in Eq.~\eqref{eq:cam_logit}, each logit becomes
\[
  \ell'_k \;=\; \sum_{j=1}^N w'_{j,k}\,\bar F_j + b_k
            \;=\; \ell_k + \delta\,\bar F_i,
\]
so softmax probabilities are unchanged by \eqref{eq:shift_invariance}.
However, the linear CAM combination in Eq.~\eqref{eq:cam} changes \emph{spatially}:
\begin{equation}
  M'_k(p,q)
  \;=\; \sum_{j=1}^N w'_{j,k}\,F_j^{(p,q)}
  \;=\; \sum_{j=1}^N w_{j,k}\,F_j^{(p,q)} \;+\; \delta\,F_i^{(p,q)}
  \;=\; M_k(p,q) + \delta F_i^{(p,q)}.
  \label{eq:shift_mask}
\end{equation}
Thus -- even with identical class probabilities -- the heatmap can be arbitrarily brightened or dimmed where $F_i$ is large (and likewise after applying ReLU when used).

\paragraph{2) Sign Collapse.}
Softmax depends only on the \emph{relative differences} among logits and ignores their absolute sign.
For example, subtracting a large constant from every FC weight in Eq.~\eqref{eq:cam_logit},  
\begin{equation}
  w'_{i,k} = w_{i,k} - \delta,\quad \delta\gg0,
  \label{eq:collapse}
\end{equation}
shifts all logits uniformly (so $y_k$ remains unchanged by the shift invariance in Eq.~\eqref{eq:shift_invariance}), yet it inverts CAM contributions: channels that were positive become inhibitory, thereby misleading the localization.

\subsection{Dual-Branch Sigmoid Head}
\label{sec:method}

To disentangle these distortions, we introduce a \emph{dual-branch sigmoid} head that decouples localization from classification.

\paragraph{Head Replication.}
Starting from a pretrained classifier, we copy its head $h$ into a new branch $\tilde{h}$ with identical architecture (GAP$+$FC or MLPs), but fresh parameters.  The sigmoid branch outputs class-wise scores:
\begin{equation}
  s_k = \sigma\bigl(\tilde{h}_k(\mathbf{F})\bigr),\quad\sigma(z)=1/(1+e^{-z}),\quad k=1,\dots,C.
  \label{eq:logits_sigmoid}
\end{equation}
The original softmax head and backbone remain frozen.

\paragraph{Binary Supervision.}
Each image is a positive sample for exactly one class and a negative sample for the other $C\!-\!1$ classes, inducing a strong imbalance (\eg, 1:999 on ImageNet).
We therefore train the sigmoid head $\tilde h$ with class-balanced binary cross‐entropy:
\begin{equation}
  \mathcal{L} = -\frac1B\sum_{n=1}^B\Bigl[\,
    \underbrace{\left(1-\tfrac1{C}\right)\log s_{y^{(n)}}}_{\text{positive}} 
    + \underbrace{\tfrac1{C}\sum_{k\neq y^{(n)}}\log(1 - s_k)}_{\text{negatives}}
  \Bigr].
  \label{eq:bce_loss}
\end{equation}

\begin{figure*}[t!]
\begin{center}
\includegraphics[width=0.9\textwidth]{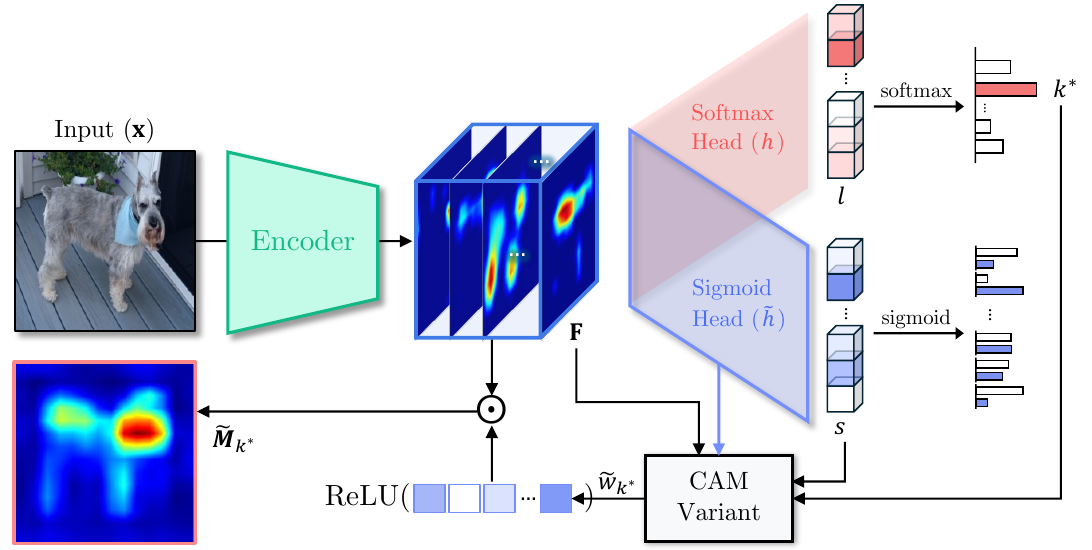}
\end{center}
\caption{
Inference pipeline of the dual‐branch sigmoid CAM.
After feature extraction, the frozen softmax head predicts the class label $k^*$.
In parallel, any CAM variant computes per‐channel importance scores $\tilde{w}_{k^*}$ (via weights or gradients) for $s_{k^*}$, which are rectified by clamping to positive values.
These positive‐only scores are then linearly combined with the feature maps to produce the final class evidence heatmap $\tilde{M}_{k^*}$.
}
\label{fig:inference}
\end{figure*}

\paragraph{Sigmoid Branch Semantics.}
Each sigmoid head $s_k=\sigma(\tilde\ell_k)$ is trained as an \emph{independent} binary classifier for class $k$ with BCE (Eq.~\eqref{eq:bce_loss}).  
Because classes are optimized independently, $\tilde w_{i,k}$ is determined solely by its effect on $s_k$ rather than by cross‐class normalization or logit ranks.
This breaks softmax invariances (additive logit shifts and sign collapse) and restores both the \emph{magnitude} and \emph{polarity} needed for CAM‐style maps.
For example, under a GAP$+$FC head (as in vanilla CAM), the sigmoid logit can be written as $\tilde\ell_k=\tilde b_k+\sum_{i=1}^N \tilde w_{i,k}\,\bar F_i$ (\cf Eq.~\eqref{eq:cam_logit}), and it satisfies:
\begin{itemize}[leftmargin=*]
    \item \textbf{Relative importance.}
    The BCE objective increases $\tilde w_{i,k}$ for channels whose activations raise $s_k$ on positives and decreases it otherwise, yielding within‐class comparability: larger $|\tilde w_{i,k}|$ denotes stronger influence on $s_k$ for class $k$.
    \item \textbf{Sign semantics (evidence‐only mapping).}
    Since $\sigma(\cdot)$ is strictly increasing, increasing $\bar F_i$ raises $s_k$ iff $\tilde w_{i,k}>0$ (evidence) and lowers it iff $\tilde w_{i,k}<0$ (inhibition), thereby restoring the polarity lost under softmax.
    Leveraging this at inference, we retain only positive contributions -- using $\max(0,\tilde w_{i,k})$ (or $\max(0,\tilde\alpha_{i,k})$) -- and combine them with feature maps as in Eq.~\eqref{eq:cam} to produce class-$k$ evidence maps without negative leakage.
\end{itemize}

\paragraph{Inference.}
During testing, we first predict the class via the softmax branch and obtain $k^*=\arg\max_k\,y_k$. 
For localization, we apply the chosen CAM variant to the \emph{sigmoid} branch -- \ie compute per-channel scores $\tilde w_{i,k^*}$ from the sigmoid logit $s_{k^*}$ rather than the softmax logit $\ell_{k^*}$ (\cf Eqs.~\eqref{eq:gradcam_importance}, \eqref{eq:score_importance}).
We then retain only positive evidence by clamping negatives to zero and combine the resulting weights with feature maps as in Eq.~\eqref{eq:relu_convention} to obtain $\tilde M_{k^*}$.
This drop-in substitution (softmax $\rightarrow$ sigmoid) makes the procedure applicable to any CAM variant and yields signed, distortion-free maps.
Figure~\ref{fig:inference} illustrates the complete inference pipeline.

\section{Experiments}
\label{sec:experiments}

We evaluate our dual-branch sigmoid head on two complementary tasks: (1) explanation fidelity in fine-grained classification, which measures how faithfully heatmaps highlight the subtle, class-specific cues that distinguish similar categories; and (2) weakly-supervised object localization (WSOL) on large-scale datasets, which tests whether our method can recover full object extents from image-level supervision.
For vanilla CAM both softmax and sigmoid heads employ a GAP+FC design; for all other variants we preserve the original classifier head architectures.
Full implementation details and additional results are provided in the Appendix.

\subsection{Fine‐Grained Explanation Fidelity}
\label{sec:exp-fg}

Fine-grained classification requires distinguishing very similar categories based on subtle visual cues.
Explanation fidelity metrics measure whether heatmaps correctly highlight these small, discriminative regions, and thus reflect the true model reasoning.

\paragraph{Datasets.}  
We evaluate two fine-grained benchmarks:
\begin{itemize}[leftmargin=*]
  \item \textbf{CUB-200-2011}~\cite{wah2011cub}: $6{,}033$ train, $5{,}755$ test images of $200$ bird species.
  \item \textbf{Stanford Cars}~\cite{krause2013cars}: $8{,}144$ train, $8{,}041$ test images of $196$ car models.
\end{itemize}

\paragraph{Metrics.}
We use masking-based \emph{Average Drop} (\%) and \emph{\% Increase in Confidence}~\cite{chattopadhay2018grad}. 
Following~\cite{chattopadhay2018grad}, the explanation map is constructed by element-wise multiplying the class-conditional saliency map (upsampled to the input size) with the original image as 
$E_{k^*} = \tilde{M}_{k^*} \circ \mathbf{x}$,
 where $\circ$ denotes the Hadamard product, $\tilde{M}_{k^*}$ is the saliency map for predicted class $k^*$, and $\mathbf{x}$ is the input image.
These quantify whether the regions highlighted by a heatmap preserve or boost the target class score, serving as proxies for explanation faithfulness.

\paragraph{Backbones \& Methods.}  
We use VGG‐16~\cite{VGG:Simonyan2014} and ResNet‐50~\cite{Resnet:He2015,SE:Hu2017} backbones.
To demonstrate generality, we evaluate five CAM variants -- CAM~\cite{zhou2016learning}, Grad‐CAM
~\cite{selvaraju2017grad}, Grad‐CAM++~\cite{chattopadhay2018grad}, XGrad‐CAM~\cite{fu2020axiom}, and Layer‐CAM~\cite{jiang2021layercam} -- each with and without our sigmoid branch.
This comprehensive setup shows how the sigmoid add‐on improves diverse explanation techniques under fine‐grained conditions, particularly for gradient-based methods.

\begin{table}[t!]
\small
\setlength{\tabcolsep}{4pt}
\begin{center}
\begin{tabular}{l|l|cc|cc}
\toprule
\multirow{2}{*}{Backbone} & \multirow{2}{*}{Method}
  & \multicolumn{2}{c|}{CUB-200-2011}
  & \multicolumn{2}{c}{Stanford Cars} \\
 & & \% Avg Drop ($\downarrow$) & \% Inc Conf ($\uparrow$)
     & \% Avg Drop ($\downarrow$) & \% Inc Conf ($\uparrow$) \\
\midrule
\multirow{10}{*}{VGG-16}
  & \cellcolor{gray!15}CAM
    & \cellcolor{gray!15}45.36 & \cellcolor{gray!15}7.08
    & \cellcolor{gray!15}31.45 & \cellcolor{gray!15}9.68 \\
  & $+$ Ours
    & 37.20 {\scriptsize\textcolor{blue}{(-8.16)}} 
    & 18.86 {\scriptsize\textcolor{blue}{(+11.78)}} 
    & 43.58 {\scriptsize\textcolor{red}{(+12.13)}} 
    & 10.74 {\scriptsize\textcolor{blue}{(+1.06)}} \\
  & \cellcolor{gray!15}Grad-CAM
    & \cellcolor{gray!15}38.88 & \cellcolor{gray!15}11.36
    & \cellcolor{gray!15}56.10 & \cellcolor{gray!15}8.11 \\
  & $+$ Ours
    & 35.66 {\scriptsize\textcolor{blue}{(-3.22)}} 
    & 22.06 {\scriptsize\textcolor{blue}{(+10.70)}} 
    &  9.96 {\scriptsize\textcolor{blue}{(-46.14)}} 
    & 19.46 {\scriptsize\textcolor{blue}{(+11.35)}} \\
  & \cellcolor{gray!15}Grad-CAM++
    & \cellcolor{gray!15}31.11 & \cellcolor{gray!15}14.46
    & \cellcolor{gray!15}39.63 & \cellcolor{gray!15}11.71 \\
  & $+$ Ours
    & 33.98 {\scriptsize\textcolor{red}{(+2.87)}} 
    & 21.75 {\scriptsize\textcolor{blue}{(+7.29)}} 
    &  9.46 {\scriptsize\textcolor{blue}{(-30.17)}} 
    & 18.46 {\scriptsize\textcolor{blue}{(+6.75)}} \\
  & \cellcolor{gray!15}XGrad-CAM
    & \cellcolor{gray!15}38.56 & \cellcolor{gray!15}11.10
    & \cellcolor{gray!15}50.59 & \cellcolor{gray!15}9.14 \\
  & $+$ Ours
    & 38.04 {\scriptsize\textcolor{blue}{(-0.52)}} 
    & 20.99 {\scriptsize\textcolor{blue}{(+9.89)}} 
    &  9.69 {\scriptsize\textcolor{blue}{(-40.90)}} 
    & 20.53 {\scriptsize\textcolor{blue}{(+11.39)}} \\
  & \cellcolor{gray!15}Layer-CAM
    & \cellcolor{gray!15}37.19 & \cellcolor{gray!15}11.27
    & \cellcolor{gray!15}47.51 & \cellcolor{gray!15}9.70 \\
  & $+$ Ours
    & 41.58 {\scriptsize\textcolor{red}{(+4.39)}} 
    & 19.50 {\scriptsize\textcolor{blue}{(+8.23)}} 
    & 10.51 {\scriptsize\textcolor{blue}{(-37.00)}} 
    & 19.39 {\scriptsize\textcolor{blue}{(+9.69)}} \\
\midrule
\multirow{10}{*}{ResNet-50}
  & \cellcolor{gray!15}CAM
    & \cellcolor{gray!15}44.17 & \cellcolor{gray!15}14.76
    & \cellcolor{gray!15}48.86 & \cellcolor{gray!15}4.32 \\
  & $+$ Ours
    & 13.86 {\scriptsize\textcolor{blue}{(-30.31)}} 
    & 18.42 {\scriptsize\textcolor{blue}{(+3.66)}} 
    &  3.77 {\scriptsize\textcolor{blue}{(-45.09)}} 
    & 40.62 {\scriptsize\textcolor{blue}{(+36.30)}} \\
  & \cellcolor{gray!15}Grad-CAM
    & \cellcolor{gray!15}46.38 & \cellcolor{gray!15}13.07
    & \cellcolor{gray!15}29.53 & \cellcolor{gray!15}14.56 \\
  & $+$ Ours
    & 19.07 {\scriptsize\textcolor{blue}{(-27.31)}} 
    & 13.22 {\scriptsize\textcolor{blue}{(+0.15)}} 
    &  3.66 {\scriptsize\textcolor{blue}{(-25.87)}} 
    & 41.41 {\scriptsize\textcolor{blue}{(+26.85)}} \\
  & \cellcolor{gray!15}Grad-CAM++
    & \cellcolor{gray!15}43.15 & \cellcolor{gray!15}14.50
    & \cellcolor{gray!15}24.68 & \cellcolor{gray!15}17.31 \\
  & $+$ Ours
    & 18.88 {\scriptsize\textcolor{blue}{(-24.27)}} 
    & 13.72 {\scriptsize\textcolor{red}{(-0.78)}} 
    &  3.65 {\scriptsize\textcolor{blue}{(-21.03)}} 
    & 41.62 {\scriptsize\textcolor{blue}{(+24.31)}} \\
  & \cellcolor{gray!15}XGrad-CAM
    & \cellcolor{gray!15}46.38 & \cellcolor{gray!15}13.07
    & \cellcolor{gray!15}29.53 & \cellcolor{gray!15}14.56 \\
  & $+$ Ours
    & 19.07 {\scriptsize\textcolor{blue}{(-27.31)}} 
    & 13.22 {\scriptsize\textcolor{blue}{(+0.15)}} 
    &  3.66 {\scriptsize\textcolor{blue}{(-25.87)}} 
    & 41.41 {\scriptsize\textcolor{blue}{(+26.85)}} \\
  & \cellcolor{gray!15}Layer-CAM
    & \cellcolor{gray!15}43.69 & \cellcolor{gray!15}13.91
    & \cellcolor{gray!15}22.64 & \cellcolor{gray!15}18.90 \\
  & $+$ Ours
    & 19.07 {\scriptsize\textcolor{blue}{(-24.62)}} 
    & 13.22 {\scriptsize\textcolor{red}{(-0.69)}} 
    &  3.66 {\scriptsize\textcolor{blue}{(-18.98)}} 
    & 41.39 {\scriptsize\textcolor{blue}{(+22.49)}} \\
\bottomrule
\end{tabular}
\end{center}
\caption{Fine-grained explanation fidelity on CUB-200-2011 and Stanford Cars. For \% Average Drop (lower is better) and \% Increase in Confidence (higher is better), improved values are shown in blue and worsened values in red; parentheses indicate the change relative to the baseline.}
\label{tab:fg_results}
\end{table}

\paragraph{Results.}
Table~\ref{tab:fg_results} demonstrates that our sigmoid branch markedly improves explanation fidelity across both backbones.
On CUB-200-2011 with VGG-16, average drop is reduced by $0.52$-$8.16$ percentage points (\eg, CAM: $45.36\to37.20$, $-8.16$) while confidence increase rises by $7.29$-$11.78$ points.
Stanford Cars shows similar trends: most methods cut average drop by $30$-$46$ points (\eg, Grad-CAM: $56.10\to9.96$, $-46.14$), and boost confidence by $1$-$11$ points.
ResNet-50 yields even larger gains -- CAM's drop falls from $44.17$ to $13.86$ ($-30.31$) and its confidence jump from $14.76$ to $18.42$ ($+3.66$), with analogous improvements for other variants.
Only Grad-CAM++ on CUB and one Layer-CAM case exhibit minor reversals; overall, our approach sharply reduces drop and elevates confidence without exception or trade-offs. 

\subsection{Weakly‐Supervised Object Localization}
\label{sec:exp-wsol}

WSOL benchmarks assess a model's ability to localize entire objects using only image-level labels.
Because traditional CAM-based explanations often focus on the most discriminative parts, WSOL on large-scale datasets reveals whether our sigmoid branch restores complete object coverage.

\paragraph{Datasets.}
We use two standard WSOL benchmarks:
\begin{itemize}[leftmargin=*]
  \item \textbf{ImageNet-1K}~\cite{Imagenet:Russakovsky2015}: $1.28M$ train, $50K$ val images over $1{,}000$ classes, with bounding boxes used only for evaluation.
  \item \textbf{OpenImages-30K}~\cite{OpenImages:Benenson2019}: $29{,}819$ train, $2{,}500$ val, and $5{,}000$ test images with pixel-wise masks.
\end{itemize}

\begin{table}[t!]
\fontsize{8.5}{10}\selectfont
\setlength{\tabcolsep}{6pt}
\begin{center}
\begin{tabular}{l|l|cccc|cc}
\toprule
\multirow{2}{*}{Backbone} & \multirow{2}{*}{Method}
  & \multicolumn{4}{c|}{ImageNet-1K}
  & \multicolumn{2}{c}{OpenImages-30K} \\
 & & Top1 Cls & Top1 Loc & GT Loc & MBAv2 & Top1 Cls & PxAP \\
\midrule
\multirow{6}{*}{VGG-16}
  & \cellcolor{gray!15}CAM             
    & \cellcolor{gray!15}66.56 & \cellcolor{gray!15}43.13 
    & \cellcolor{gray!15}59.54 & \cellcolor{gray!15}60.00 
    & \cellcolor{gray!15}70.00 & \cellcolor{gray!15}58.17 \\
  & $+$ Ours      
    & --    & 43.14 
    & 59.57
    & 60.25
    & --    & 60.11\\
  & $\Delta$        
    & --    & {\scriptsize\textcolor{blue}{(+0.01)}} 
    & {\scriptsize\textcolor{blue}{(+0.03)}} 
    & {\scriptsize\textcolor{blue}{(+0.25)}} 
    & --    & {\scriptsize\textcolor{blue}{(+1.94)}} \\
  & \cellcolor{gray!15}Grad-CAM        
    & \cellcolor{gray!15}69.61 & \cellcolor{gray!15}37.04 
    & \cellcolor{gray!15}49.02 & \cellcolor{gray!15}52.49 
    & \cellcolor{gray!15}70.12 & \cellcolor{gray!15}55.25 \\
  & $+$ Ours 
    & --    & 44.68
    & 59.43
    & 60.17
    & --    & 56.53 \\
  & $\Delta$        
    & --    & {\scriptsize\textcolor{blue}{(+7.64)}} 
    & {\scriptsize\textcolor{blue}{(+10.41)}} 
    & {\scriptsize\textcolor{blue}{(+7.68)}} 
    & --    & {\scriptsize\textcolor{blue}{(+1.28)}} \\
\midrule
\multirow{6}{*}{ResNet-50}
  & \cellcolor{gray!15}CAM             
    & \cellcolor{gray!15}73.89 & \cellcolor{gray!15}47.69 
    & \cellcolor{gray!15}60.80 & \cellcolor{gray!15}63.53 
    & \cellcolor{gray!15}74.02 & \cellcolor{gray!15}59.36 \\
  & $+$ Ours      
    & --    & 49.44
    & 62.68
    & 64.73
    & --    & 60.49\\
  & $\Delta$        
    & --    & {\scriptsize\textcolor{blue}{(+1.75)}}
    & {\scriptsize\textcolor{blue}{(+1.88)}}
    & {\scriptsize\textcolor{blue}{(+1.20)}}
    & --    & {\scriptsize\textcolor{blue}{(+1.13)}} \\
  & \cellcolor{gray!15}Grad-CAM        
    & \cellcolor{gray!15}69.01 & \cellcolor{gray!15}41.14 
    & \cellcolor{gray!15}55.28 & \cellcolor{gray!15}57.04 
    & \cellcolor{gray!15}74.02 & \cellcolor{gray!15}60.56 \\
  & $+$ Ours 
    & --    & 40.34
    & 53.61
    & 57.98
    & --    & 60.24 \\
  & $\Delta$        
    & --    & {\scriptsize\textcolor{red}{(-0.80)}}
    & {\scriptsize\textcolor{red}{(-1.67)}}
    & {\scriptsize\textcolor{blue}{(+0.94)}}
    & --    & {\scriptsize\textcolor{red}{(-0.32)}} \\
\midrule
\multirow{6}{*}{InceptionV3}
  & \cellcolor{gray!15}CAM             
    & \cellcolor{gray!15}69.72 & \cellcolor{gray!15}42.16 
    & \cellcolor{gray!15}57.69 & \cellcolor{gray!15}63.51 
    & \cellcolor{gray!15}56.22 & \cellcolor{gray!15}62.25 \\
  & $+$ Ours      
    & --    & 44.11
    & 60.12
    & 64.98
    & --    & 63.65\\
  & $\Delta$        
    & --    & {\scriptsize\textcolor{blue}{(+1.95)}}
    & {\scriptsize\textcolor{blue}{(+2.43)}}
    & {\scriptsize\textcolor{blue}{(+1.47)}}
    & --    & {\scriptsize\textcolor{blue}{(+1.40)}} \\
  & \cellcolor{gray!15}Grad-CAM        
    & \cellcolor{gray!15}67.79 & \cellcolor{gray!15}34.57 
    & \cellcolor{gray!15}46.35 & \cellcolor{gray!15}52.30 
    & \cellcolor{gray!15}68.56 & \cellcolor{gray!15}49.64 \\
  & $+$ Ours 
    & --    & 33.72
    & 45.18
    & 53.06
    & --    & 48.16\\
  & $\Delta$        
    & --    & {\scriptsize\textcolor{red}{(-0.85)}}
    & {\scriptsize\textcolor{red}{(-1.17)}}
    & {\scriptsize\textcolor{blue}{(+0.76)}}
    & --    & {\scriptsize\textcolor{red}{(-1.48)}} \\
\bottomrule
\end{tabular}
\end{center}
\caption{WSOL results on ImageNet-1K and OpenImages-30K. For each base method we shade the baseline row in gray; ``$+$ Ours'' rows report updated scores with their Δ shown in parentheses (blue for gains, red for drops).}
\label{tab:wsol_results}
\end{table}

\paragraph{Metrics.}
On ImageNet-1K, we report Top-1 classification accuracy, Top-1 localization, GT-known localization (IoU$\geq 0.5$), and MaxBoxAccV2 (MBAv2)~\cite{EvalWSOL:Choe2019}.
On OpenImages-30K, we report Top-1 classification and pixel-level average precision (PxAP) over the masks~\cite{EvalWSOL:Choe2019}.
These metrics together ensure that our sigmoid branch maintains recognition performance while improving localization.

\paragraph{Backbones \& Methods.}  
We test three widely‐used architectures: VGG‐16~\cite{VGG:Simonyan2014}, ResNet‐50~\cite{Resnet:He2015,SE:Hu2017}, and InceptionV3~\cite{Inception:Szegedy2014}.
As base localization methods, we choose CAM~\cite{zhou2016learning} and Grad‐CAM~\cite{selvaraju2017grad}, then integrate our sigmoid branch into each.
This setup evaluates how well our add‐on improves diverse architectures and CAM variants.

\paragraph{Results.}
Table~\ref{tab:wsol_results} summarizes our WSOL performance.
With VGG-16, CAM + Ours yields slight gains in Top-1 Loc ($+0.01$), GT Loc ($+0.03$), MBAv2 ($+0.25$) and a larger boost in PxAP ($+1.94$), while Grad-CAM + Ours delivers substantial improvements ($+7.64$ Top-1 Loc, $+10.41$ GT Loc, $+7.68$ MBAv2, $+1.28$ PxAP).
For ResNet-50, CAM + Ours improves all metrics, whereas Grad-CAM + Ours trades minor drops in Top-1 Loc ($-0.80$), GT Loc ($-1.67$) and PxAP ($-0.32$) for a MBAv2 gain ($+0.94$).
InceptionV3 + CAM + Ours shows consistent gains, and InceptionV3 + Grad-CAM + Ours experiences small declines in Top-1 Loc ($-0.85$), GT Loc ($-1.17$) and PxAP ($-1.48$) alongside a $+0.76$ MBAv2 increase.
Overall, our sigmoid branch consistently enhances CAM-based localization, with only minor trade-offs in select Grad-CAM cases. 

\subsection{Qualitative Results}
\label{sec:qualitative}

\begin{figure*}[htbp]
\begin{center}
\includegraphics[width=0.95\textwidth]{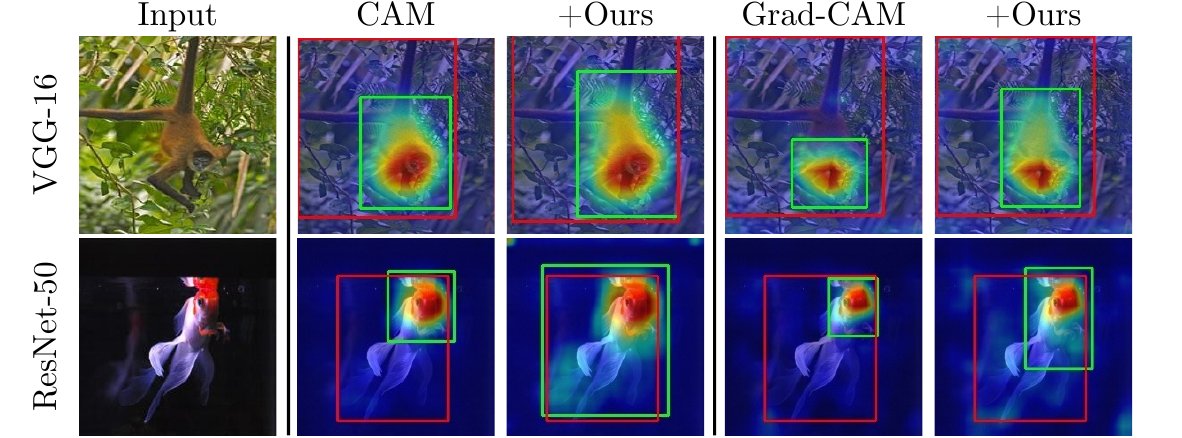}
\end{center}
\caption{Qualitative WSOL on ImageNet-1K for VGG-16 (top) and ResNet-50 (bottom). From left to right: input, CAM, CAM$+$Ours, Grad-CAM, Grad-CAM$+$Ours. The boxes in green and red represent the predictions and ground truths of localization.
}
\label{fig:qual_results}
\end{figure*}

Figure~\ref{fig:qual_results} shows two ImageNet WSOL examples -- one with VGG-16 and one with ResNet-50. 
Our sigmoid branch consistently extends heatmaps to cover the full object rather than just the most discriminative part. 
Additional qualitative results are provided in the Appendix.

\section{Conclusion}
We have identified two inherent flaws -- \emph{additive logit shift} and \emph{sign collapse} -- in all softmax-based CAM explanations, and shown how they can lead to misleading localization.
To remedy this, we introduced a dual-branch sigmoid head that restores absolute and signed feature importance without touching the backbone or softmax branch.
Our approach is architecture-agnostic and compatible with most CAM variants.
Extensive experiments on fine‐grained explanation and WSOL benchmarks demonstrate that our method consistently improves explanation fidelity and localization robustness, all without degrading classification accuracy.

\newpage

\section*{Acknowledgements}
We thank Wonho Bae and Jinhwan Seo for insightful early discussions that helped us refine and articulate the initial problem statement of this work.
This work was supported by Institute of Information \& communications Technology Planning \& Evaluation (IITP) grant funded by the Korea government (MSIT) (No. RS-2022-00155966, Artificial Intelligence Convergence Innovation Human Resources Development (Ewha Womans University)).

\bibliography{egbib}

\begin{thebibliography}{32}
\providecommand{\natexlab}[1]{#1}
\providecommand{\url}[1]{\texttt{#1}}
\expandafter\ifx\csname urlstyle\endcsname\relax
  \providecommand{\doi}[1]{doi: #1}\else
  \providecommand{\doi}{doi: \begingroup \urlstyle{rm}\Url}\fi

\bibitem[Abnar and Zuidema(2020)]{Rollout:Abnar2020}
Samira Abnar and Willem Zuidema.
\newblock Quantifying attention flow in transformers.
\newblock \emph{arXiv preprint arXiv:2005.00928}, 2020.

\bibitem[Bach et~al.(2015)Bach, Binder, Montavon, Klauschen, M{\"u}ller, and Samek]{bach2015pixel}
Sebastian Bach, Alexander Binder, Gr{\'e}goire Montavon, Frederick Klauschen, Klaus-Robert M{\"u}ller, and Wojciech Samek.
\newblock On pixel-wise explanations for non-linear classifier decisions by layer-wise relevance propagation.
\newblock \emph{PloS one}, 10\penalty0 (7):\penalty0 e0130140, 2015.

\bibitem[Bae et~al.(2020)Bae, Noh, and Kim]{bae2020rethinking}
Wonho Bae, Junhyug Noh, and Gunhee Kim.
\newblock Rethinking class activation mapping for weakly supervised object localization.
\newblock In \emph{Computer Vision--ECCV 2020: 16th European Conference, Glasgow, UK, August 23--28, 2020, Proceedings, Part XV 16}, pages 618--634. Springer, 2020.

\bibitem[Benenson et~al.(2019)Benenson, Popov, and Ferrari]{OpenImages:Benenson2019}
Rodrigo Benenson, Stefan Popov, and Vittorio Ferrari.
\newblock {Large-scale Interactive Object Segmentation with Human Annotators}.
\newblock In \emph{CVPR}, 2019.

\bibitem[Chattopadhay et~al.(2018)Chattopadhay, Sarkar, Howlader, and Balasubramanian]{chattopadhay2018grad}
Aditya Chattopadhay, Anirban Sarkar, Prantik Howlader, and Vineeth~N Balasubramanian.
\newblock Grad-cam++: Generalized gradient-based visual explanations for deep convolutional networks.
\newblock In \emph{2018 IEEE winter conference on applications of computer vision (WACV)}, pages 839--847. IEEE, 2018.

\bibitem[Choe and Shim(2019)]{ADL:Choe2019}
Junsuk Choe and Hyunjung Shim.
\newblock {Attention-Based Dropout Layer for Weakly Supervised Object Localization}.
\newblock In \emph{CVPR}, 2019.

\bibitem[Choe et~al.(2020)Choe, Oh, Lee, Chun, Akata, and Shim]{EvalWSOL:Choe2019}
Junsuk Choe, Seong~Joon Oh, Seungho Lee, Sanghyuk Chun, Zeynep Akata, and Hyunjung Shim.
\newblock {Evaluating Weakly Supervised Object Localization Methods Right}.
\newblock In \emph{CVPR}, 2020.

\bibitem[Fong et~al.(2019)Fong, Patrick, and Vedaldi]{fong2019understanding}
Ruth Fong, Mandela Patrick, and Andrea Vedaldi.
\newblock Understanding deep networks via extremal perturbations and smooth masks.
\newblock In \emph{Proceedings of the IEEE/CVF international conference on computer vision}, pages 2950--2958, 2019.

\bibitem[Fu et~al.(2020)Fu, Hu, Dong, Guo, Gao, and Li]{fu2020axiom}
Ruigang Fu, Qingyong Hu, Xiaohu Dong, Yulan Guo, Yinghui Gao, and Biao Li.
\newblock Axiom-based grad-cam: Towards accurate visualization and explanation of cnns.
\newblock \emph{arXiv preprint arXiv:2008.02312}, 2020.

\bibitem[Gao et~al.(2021)Gao, Wan, Pan, Peng, Tian, Han, Zhou, and Ye]{TSCAM:Gao2021}
Wei Gao, Fang Wan, Xingjia Pan, Zhiliang Peng, Qi~Tian, Zhenjun Han, Bolei Zhou, and Qixiang Ye.
\newblock Ts-cam: Token semantic coupled attention map for weakly supervised object localization.
\newblock In \emph{Proceedings of the IEEE/CVF international conference on computer vision}, pages 2886--2895, 2021.

\bibitem[He et~al.(2016)He, Zhang, Ren, and Sun]{Resnet:He2015}
Kaiming He, Xiangyu Zhang, Shaoqing Ren, and Jian Sun.
\newblock {Deep Residual Learning for Image Recognition}.
\newblock In \emph{CVPR}, 2016.

\bibitem[Hu et~al.(2018)Hu, Shen, and Sun]{SE:Hu2017}
Jie Hu, Li~Shen, and Gang Sun.
\newblock {Squeeze-and-Excitation Networks}.
\newblock In \emph{CVPR}, 2018.

\bibitem[Jiang et~al.(2021)Jiang, Zhang, Hou, Cheng, and Wei]{jiang2021layercam}
Peng-Tao Jiang, Chang-Bin Zhang, Qibin Hou, Ming-Ming Cheng, and Yunchao Wei.
\newblock Layercam: Exploring hierarchical class activation maps for localization.
\newblock \emph{IEEE Transactions on Image Processing}, 30:\penalty0 5875--5888, 2021.

\bibitem[Krause et~al.(2013)Krause, Stark, Deng, and Fei-Fei]{krause2013cars}
Jonathan Krause, Michael Stark, Jia Deng, and Li~Fei-Fei.
\newblock 3d object representations for fine-grained categorization.
\newblock In \emph{Proceedings of the IEEE international conference on computer vision workshops}, pages 554--561, 2013.

\bibitem[Muhammad and Yeasin(2020)]{muhammad2020eigen}
Mohammed~Bany Muhammad and Mohammed Yeasin.
\newblock Eigen-cam: Class activation map using principal components.
\newblock In \emph{2020 international joint conference on neural networks (IJCNN)}, pages 1--7. IEEE, 2020.

\bibitem[Petsiuk et~al.(2018)Petsiuk, Das, and Saenko]{petsiuk2018rise}
Vitali Petsiuk, Abir Das, and Kate Saenko.
\newblock Rise: Randomized input sampling for explanation of black-box models.
\newblock \emph{arXiv preprint arXiv:1806.07421}, 2018.

\bibitem[Ramaswamy et~al.(2020)]{ramaswamy2020ablation}
Harish~Guruprasad Ramaswamy et~al.
\newblock Ablation-cam: Visual explanations for deep convolutional network via gradient-free localization.
\newblock In \emph{proceedings of the IEEE/CVF winter conference on applications of computer vision}, pages 983--991, 2020.

\bibitem[Russakovsky et~al.(2015)Russakovsky, Deng, SU, Krause, Satheesh, Ma, Huang, Karpathy, Khosla, Bernstein, Berg, and Fei-Fei]{Imagenet:Russakovsky2015}
Olga Russakovsky, Jia Deng, Hao SU, Jonathan Krause, Sanjeev Satheesh, Sean Ma, Zhiheng Huang, Andrej Karpathy, Aditya Khosla, Michael Bernstein, Alexander~C. Berg, and LI~Fei-Fei.
\newblock {Imagenet Large Scale Visual Recognition Challenge}.
\newblock \emph{IJCV}, 2015.

\bibitem[Selvaraju et~al.(2017)Selvaraju, Cogswell, Das, Vedantam, Parikh, and Batra]{selvaraju2017grad}
Ramprasaath~R Selvaraju, Michael Cogswell, Abhishek Das, Ramakrishna Vedantam, Devi Parikh, and Dhruv Batra.
\newblock Grad-cam: Visual explanations from deep networks via gradient-based localization.
\newblock In \emph{Proceedings of the IEEE international conference on computer vision}, pages 618--626, 2017.

\bibitem[Shrikumar et~al.(2017)Shrikumar, Greenside, and Kundaje]{shrikumar2017learning}
Avanti Shrikumar, Peyton Greenside, and Anshul Kundaje.
\newblock Learning important features through propagating activation differences.
\newblock In \emph{International conference on machine learning}, pages 3145--3153. PMlR, 2017.

\bibitem[Simonyan and Zisserman(2014)]{VGG:Simonyan2014}
Karen Simonyan and Andrew Zisserman.
\newblock {Very Deep Convolutional Networks for Large-Scale Image Recognition}.
\newblock \emph{Arxiv:1409.1556}, 2014.

\bibitem[Simonyan et~al.(2013)Simonyan, Vedaldi, and Zisserman]{simonyan2013deep}
Karen Simonyan, Andrea Vedaldi, and Andrew Zisserman.
\newblock Deep inside convolutional networks: Visualising image classification models and saliency maps.
\newblock \emph{arXiv preprint arXiv:1312.6034}, 2013.

\bibitem[Singh and Lee(2017)]{HaS:Singh2017}
Krishna~Kumar Singh and Yong~Jae Lee.
\newblock {Hide-And-Seek: Forcing a Network to Be Meticulous for Weakly-Supervised Object and Action Localization}.
\newblock In \emph{ICCV}, 2017.

\bibitem[Smilkov et~al.(2017)Smilkov, Thorat, Kim, Vi{\'e}gas, and Wattenberg]{smilkov2017smoothgrad}
Daniel Smilkov, Nikhil Thorat, Been Kim, Fernanda Vi{\'e}gas, and Martin Wattenberg.
\newblock Smoothgrad: removing noise by adding noise.
\newblock \emph{arXiv preprint arXiv:1706.03825}, 2017.

\bibitem[Sundararajan et~al.(2017)Sundararajan, Taly, and Yan]{sundararajan2017axiomatic}
Mukund Sundararajan, Ankur Taly, and Qiqi Yan.
\newblock Axiomatic attribution for deep networks.
\newblock In \emph{International conference on machine learning}, pages 3319--3328. PMLR, 2017.

\bibitem[Szegedy et~al.(2015)Szegedy, Liu, Jia, Sermanet, Reed, Anguelov, Erhan, Vanhoucke, and Rabinovich]{Inception:Szegedy2014}
Christian Szegedy, Wei Liu, Yangqing Jia, Pierre Sermanet, Scott Reed, Dragomir Anguelov, Dumitru Erhan, Vincent Vanhoucke, and Andrew Rabinovich.
\newblock {Going Deeper with Convolutions}.
\newblock In \emph{CVPR}, 2015.

\bibitem[Wah et~al.(2011)Wah, Branson, Welinder, Perona, and Belongie]{wah2011cub}
Catherine Wah, Steve Branson, Peter Welinder, Pietro Perona, and Serge Belongie.
\newblock The caltech-ucsd birds-200-2011 dataset.
\newblock 2011.

\bibitem[Wang et~al.(2020)Wang, Wang, Du, Yang, Zhang, Ding, Mardziel, and Hu]{wang2020score}
Haofan Wang, Zifan Wang, Mengnan Du, Fan Yang, Zijian Zhang, Sirui Ding, Piotr Mardziel, and Xia Hu.
\newblock Score-cam: Score-weighted visual explanations for convolutional neural networks.
\newblock In \emph{Proceedings of the IEEE/CVF conference on computer vision and pattern recognition workshops}, pages 24--25, 2020.

\bibitem[Wei et~al.(2017)Wei, Feng, Liang, Cheng, Zhao, and Yan]{Wei2017}
Yunchao Wei, Jiashi Feng, Xiaodan Liang, Ming-Ming Cheng, Yao Zhao, and Shuicheng Yan.
\newblock {Object Region Mining With Adversarial Erasing: A Simple Classification to Semantic Segmentation Approach}.
\newblock In \emph{CVPR}, 2017.

\bibitem[Zhang et~al.(2018)Zhang, Wei, Feng, Yang, and Huang]{ACoL:Zhang2018}
Xiaolin Zhang, Yunchao Wei, Jiashi Feng, Yi~Yang, and Thomas~S Huang.
\newblock {Adversarial Complementary Learning for Weakly Supervised Object Localization}.
\newblock In \emph{CVPR}, 2018.

\bibitem[Zhang et~al.(2025)Zhang, Gu, Chowdhury, Mai, Carlyn, Berger-Wolf, Su, and Chao]{zhang2025finer}
Ziheng Zhang, Jianyang Gu, Arpita Chowdhury, Zheda Mai, David Carlyn, Tanya Berger-Wolf, Yu~Su, and Wei-Lun Chao.
\newblock Finer-cam: Spotting the difference reveals finer details for visual explanation.
\newblock \emph{arXiv preprint arXiv:2501.11309}, 2025.

\bibitem[Zhou et~al.(2016)Zhou, Khosla, Lapedriza, Oliva, and Torralba]{zhou2016learning}
Bolei Zhou, Aditya Khosla, Agata Lapedriza, Aude Oliva, and Antonio Torralba.
\newblock Learning deep features for discriminative localization.
\newblock In \emph{Proceedings of the IEEE conference on computer vision and pattern recognition}, pages 2921--2929, 2016.

\end{thebibliography}

\clearpage
\appendix

\section{Computational Overhead}  
We quantify the additional cost of our method in terms of training time, inference latency, and parameter count.  
Inference time is measured \emph{per image} and includes the cost of CAM generation.  
All measurements are taken on ImageNet-1K using a single NVIDIA RTX A6000.

\vspace{-10pt}
\paragraph{Training time.}
As summarized in Table~\ref{tab:overhead_all}, the training overhead is moderate: \(\,+4.80\%\) for InceptionV3 (50.42\,h \(\to\) 52.84\,h), \(+18.11\%\) for VGG-16 (17.67\,h \(\to\) 20.87\,h), and \(+21.48\%\) for ResNet-50 (58.19\,h \(\to\) 70.69\,h).  
This increase mainly stems from optimizing an additional (lightweight) sigmoid head with BCE while sharing the backbone feature extractor; the backbone remains frozen and no extra convolutional layers are introduced.

\vspace{-10pt}
\paragraph{Inference time.}
Per-image latency shows a similar, small uptick: \(+14.15\%\) for InceptionV3 (108.1\,ms \(\to\) 123.4\,ms), \(+17.89\%\) for ResNet-50 (117.4\,ms \(\to\) 138.4\,ms), and \(+19.48\%\) for VGG-16 (84.7\,ms \(\to\) 101.2\,ms).  
At test time, feature extraction is unchanged; the overhead comes from an extra pass through the duplicated classifier head and using the sigmoid branch to form the CAM (the Grad-CAM/CAM backprop cost is comparable to the baseline).

\vspace{-10pt}
\paragraph{Parameter count.}
Model size increases only by duplicating the final classifier: \(+3.89\%\) for InceptionV3 (26.51M \(\to\) 27.54M), \(+5.03\%\) for VGG-16 (20.46M \(\to\) 21.49M), and \(+8.02\%\) for ResNet-50 (25.56M \(\to\) 27.61M).  
This reflects adding \(\mathcal{O}(C\times D)\) weights of the linear head (\eg \(2048\times1000\) for ResNet-50), while the backbone is unchanged.

\medskip
Overall, our dual-branch design adds a small, well-bounded overhead while yielding the consistent WSOL and fine-grained gains reported in Secs.~\ref{sec:exp-fg}–\ref{sec:exp-wsol}.
Because the backbone is frozen and reused, the method remains practical and easily deployable in existing CAM pipelines.

\begin{table}[h]
\small
\setlength{\tabcolsep}{4pt}
\centering
\resizebox{\textwidth}{!}{%
\begin{tabular}{l|ccc|ccc|ccc}
\toprule
\multirow{2}{*}{Backbone} &
\multicolumn{3}{c|}{Training Time (h)} &
\multicolumn{3}{c|}{Inference Time (ms)} &
\multicolumn{3}{c}{\# Params (M)} \\
& Baseline & Ours & $\Delta$ (\%) & Baseline & Ours & $\Delta$ (\%) & Baseline & Ours & $\Delta$ (\%) \\
\midrule
VGG-16      & 17.67 & 20.87 & +3.20 (18.11) &  84.7 & 101.2 & +16.5 (19.48) & 20.46 & 21.49 & +1.03 (5.03) \\
ResNet-50   & 58.19 & 70.69 & +12.50 (21.48) & 117.4 & 138.4 & +21.0 (17.89) & 25.56 & 27.61 & +2.05 (8.02) \\
InceptionV3 & 50.42 & 52.84 & +2.42 (4.80)  & 108.1 & 123.4 & +15.3 (14.15) & 26.51 & 27.54 & +1.03 (3.89) \\
\bottomrule
\end{tabular}%
}
\vspace{10pt}
\caption{Computational overhead on ImageNet‐1K. $\Delta$ reports the absolute change with percentage overhead in parentheses.}
\label{tab:overhead_all}
\end{table}

\section{Ablation Studies}
To isolate the contribution of each design choice, we conduct controlled studies on two axes:
(i) \emph{negative-weight clamping} (NWC) when forming CAM-style maps, and
(ii) positive-class reweighting for label imbalance in the sigmoid loss.
Here we report the first (NWC); the second is discussed in the following subsection.

\paragraph{Negative-Weight Clamping (NWC).}
Table~\ref{tab:clamping_effect} compares our sigmoid-branch maps when we \emph{keep signed weights} (Ours, w/o NWC) versus \emph{clamp negatives to zero} before heatmap composition (Ours, w/ NWC).
Across VGG-16 and ResNet-50, clamping generally improves Top-1 Loc, GT Loc, and MBAv2 for both CAM and Grad-CAM.
Even for ResNet-50 with Grad-CAM, where Top-1/GT Loc slightly drop, MBAv2 increases, indicating better spatial coverage.

\begin{table}[t!]
\centering
\fontsize{8.5}{10}\selectfont
\setlength{\tabcolsep}{6pt}
\begin{tabular}{l|l|cccc}
\toprule
Backbone & Method & Top1 Cls & Top1 Loc & GT Loc & MBAv2 \\
\midrule
\multirow{6}{*}{VGG-16}
& \cellcolor{gray!15}CAM                 & \cellcolor{gray!15}66.56 & \cellcolor{gray!15}43.13 & \cellcolor{gray!15}59.54 & \cellcolor{gray!15}60.00 \\
& Ours (w/o NWC)                          & --    & 40.12 & 53.92 & 55.24 \\
& Ours (w/ NWC)                            & --    & \textbf{43.14} & \textbf{59.57} & \textbf{60.25} \\
& \cellcolor{gray!15}Grad-CAM             & \cellcolor{gray!15}69.61 & \cellcolor{gray!15}37.04 & \cellcolor{gray!15}49.02 & \cellcolor{gray!15}52.49 \\
& Ours (w/o NWC)                          & --    & 43.24 & 56.90 & 58.45 \\
& Ours (w/ NWC)                            & --    & \textbf{44.68} & \textbf{59.43} & \textbf{60.17} \\
\midrule
\multirow{6}{*}{ResNet-50}
& \cellcolor{gray!15}CAM                  & \cellcolor{gray!15}73.89 & \cellcolor{gray!15}47.69 & \cellcolor{gray!15}60.80 & \cellcolor{gray!15}63.53 \\
& Ours (w/o NWC)                          & --    & 39.82 & 50.22 & 55.78 \\
& Ours (w/ NWC)                            & --    & \textbf{49.44} & \textbf{62.68} & \textbf{64.73} \\
& \cellcolor{gray!15}Grad-CAM             & \cellcolor{gray!15}69.01 & \cellcolor{gray!15}\textbf{41.14} & \cellcolor{gray!15}\textbf{55.28} & \cellcolor{gray!15}57.04 \\
& Ours (w/o NWC)                          & --    & 41.11 & 54.60 & 56.31 \\
& Ours (w/ NWC)                            & --    & 40.34 & 53.61 & \textbf{57.98} \\
\bottomrule
\end{tabular}
\vspace{10pt}
\caption{Ablation on \textbf{negative-weight clamping} (NWC) on ImageNet-1K. 
Both ``Ours'' rows use our sigmoid branch; \emph{w/o NWC} keeps signed channel weights, while \emph{w/ NWC} clamps negative weights to zero before map composition.
Baselines are shaded. Top-1 classification is unchanged (\texttt{--}) because the softmax head is frozen.}
\label{tab:clamping_effect}
\end{table}

\paragraph{Positive-Class Loss Weighting (BCE).}
We study the effect of the positive:negative weighting in Eq.~\eqref{eq:bce_loss} on ImageNet-1K (where $C{=}1000$). 
We sweep three settings for the positive term: \textbf{1:1} (no reweighting), \textbf{499:1} (half of $C{-}1$), and \textbf{999:1} ($C{-}1$). 
As summarized in Table~\ref{tab:pos_weight}, \emph{unweighted} training (1:1) is unstable for gradient-based explanations (Grad-CAM), causing severe localization degradation, whereas \emph{positively reweighted} losses (499:1 or 999:1) recover strong performance across backbones. 
For vanilla CAM, the three settings are comparatively close, but reweighting remains competitive and more robust across models. 
In all subsequent experiments, we default to $C{-}1$ for robustness.

\begin{table}[t!]
\centering
\fontsize{8.5}{10}\selectfont
\setlength{\tabcolsep}{6pt}
\begin{tabular}{l|l|cccc}
\toprule
Backbone & Method & Top1 Cls & Top1 Loc & GT Loc & MBAv2 \\
\midrule
\multirow{8}{*}{VGG-16}
& \cellcolor{gray!15}CAM (baseline)    & \cellcolor{gray!15}66.56 & \cellcolor{gray!15}43.13 & \cellcolor{gray!15}59.54 & \cellcolor{gray!15}60.00 \\
& Ours (1:1)                        & -- & \textbf{43.60} & \textbf{60.70} & \textbf{61.34} \\
& Ours (499:1)                      & -- & 43.19 & 59.57 & 60.28 \\
& Ours (999:1)                      & -- & 43.14 & 59.57 & 60.25 \\
& \cellcolor{gray!15}Grad-CAM (baseline)& \cellcolor{gray!15}69.61 & \cellcolor{gray!15}37.04 & \cellcolor{gray!15}49.02 & \cellcolor{gray!15}52.49 \\
& Ours (1:1)                        & -- & 9.28 & 12.45 & 44.17 \\
& Ours (499:1)                      & -- & \textbf{44.97} & \textbf{59.79} & \textbf{60.76} \\
& Ours (999:1)                      & -- & 44.68 & 59.43 & 60.17 \\
\midrule
\multirow{8}{*}{ResNet-50}
& \cellcolor{gray!15}CAM (baseline)     & \cellcolor{gray!15}73.89 & \cellcolor{gray!15}47.69 & \cellcolor{gray!15}60.80 & \cellcolor{gray!15}63.53 \\
& Ours (1:1)                        & -- & 47.29 & 59.84 & 61.54 \\
& Ours (499:1)                      & -- & 49.27 & 62.36 & 64.59 \\
& Ours (999:1)                      & -- & \textbf{49.44} & \textbf{62.68} & \textbf{64.73} \\
& \cellcolor{gray!15}Grad-CAM (baseline)& \cellcolor{gray!15}69.01 & \cellcolor{gray!15}41.14 & \cellcolor{gray!15}55.28 & \cellcolor{gray!15}57.04 \\
& Ours (1:1)                        & -- & 35.66 & 47.17 & 54.77 \\
& Ours (499:1)                      & -- & \textbf{40.55} & \textbf{53.90} & \textbf{58.05} \\
& Ours (999:1)                      & -- & 40.34 & 53.61 & 57.98 \\
\bottomrule
\end{tabular}
\vspace{10pt}
\caption{Effect of \textbf{positive-class loss weighting} in BCE on ImageNet-1K. 
Rows under each shaded baseline are our sigmoid-branch variants trained with the indicated positive:negative weight. 
``1:1'' corresponds to no reweighting; ``999:1'' equals $C{-}1$ for ImageNet-1K ($C{=}1000$). 
Top-1 classification is produced by the frozen softmax head (hence ``--'').}
\label{tab:pos_weight}
\end{table}

\section{Implementation Details}

We build our dual‐branch sigmoid head in PyTorch, extending the WSOL evaluation codebase of Choe \etal~\cite{EvalWSOL:Choe2019}\footnote{\url{https://github.com/clovaai/wsolevaluation}} and the \texttt{pytorch‐grad‐cam}\footnote{\url{https://github.com/jacobgil/pytorch-grad-cam}} repository.

The original softmax branch and backbone are frozen; only the sigmoid branch is fine‐tuned.

\paragraph{Model \& Initialization.}  
For VGG16, the sigmoid head mirrors the original \texttt{classifier}; for ResNet-50 and InceptionV3, it is a single fully‐connected layer.
We initialize weights using Kaiming and Xavier schemes (biases with truncated normal) per PyTorch defaults.

\paragraph{Loss.}  
We use PyTorch's \texttt{BCEWithLogitsLoss} with a positive‐class weight of \((C-1)\) (\eg, $999$ for ImageNet-1K, $199$ for OpenImages-30K) to counter the $1$:\(C-1\) imbalance.

\paragraph{Target Layers.}  
\begin{itemize}[leftmargin=*]
  \item VGG-16: \texttt{model.features[-1]}  
  \item ResNet-50: \texttt{model.layer4[-1]}  
  \item InceptionV3: \texttt{model.Mixed\_7c}  
\end{itemize}

\paragraph{Data Preprocessing \& Augmentation.}  
All inputs are resized to \(224\times224\) and normalize with mean \([0.485,0.456,0.406]\) and standard deviation \([0.229,0.224,0.225]\) both during training and at inference.
During training, we additionally apply random cropping and horizontal flipping prior to normalization. 

\paragraph{Optimizer \& Schedules.}  
We follow Choe \etal~\cite{EvalWSOL:Choe2019} for weight decay and overall epochs.  The sigmoid branch is fine‐tuned with Adam, exploring learning rates in \([5\mathrm{e}{-5},\,3\mathrm{e}{-2}]\) for stable convergence.

\paragraph{Training Details (WSOL).}  
\begin{itemize}
  \item Fine‐tune up to $10$ epochs on ImageNet-1K and OpenImages-30K.
  \item Batch size $32$ for all ``$+$ Ours'' runs.
  \item Learning rates:
    \begin{itemize}
      \item CAM $+$ Ours: $3e-3$ (VGG16), $1e-4$ (ResNet-50), $5e-4$ (InceptionV3).
      \item Grad-CAM $+$ Ours: $1e-4$ (VGG16), $5e-3$ (ResNet-50), $5e-4$ (InceptionV3).
    \end{itemize}
\end{itemize}

\paragraph{Training Details (Fine-Grained).}  
\begin{itemize}
  \item Fine‐tune up to $12$ epochs on CUB-200-2011 and $10$ epochs on Stanford Cars.
  \item Use the same optimizer, augmentation, batch size, and learning‐rate ranges as in WSOL.
\end{itemize}

\paragraph{Localization Evaluation.}  
For GT-known localization we threshold heatmaps at $0.2$ and use IoU$\geq0.5$.  MaxBoxAccV2 is averaged over IoU thresholds \(\{0.3,0.5,0.7\}\) with a localization threshold sweep in $0.001$ steps.

\paragraph{Layer-CAM Exception.}  
Since Layer-CAM already applies ReLU to per‐pixel importance, we omit negative‐weight clamping for those experiments.

\section{More Qualitative Results}
\label{sec:more_qual}

In this section, we provide additional qualitative examples illustrating the benefits of our dual‐branch sigmoid head for both WSOL and fine‐grained explanation tasks.

\begin{figure*}[htbp]
\begin{center}
\includegraphics[width=\textwidth]{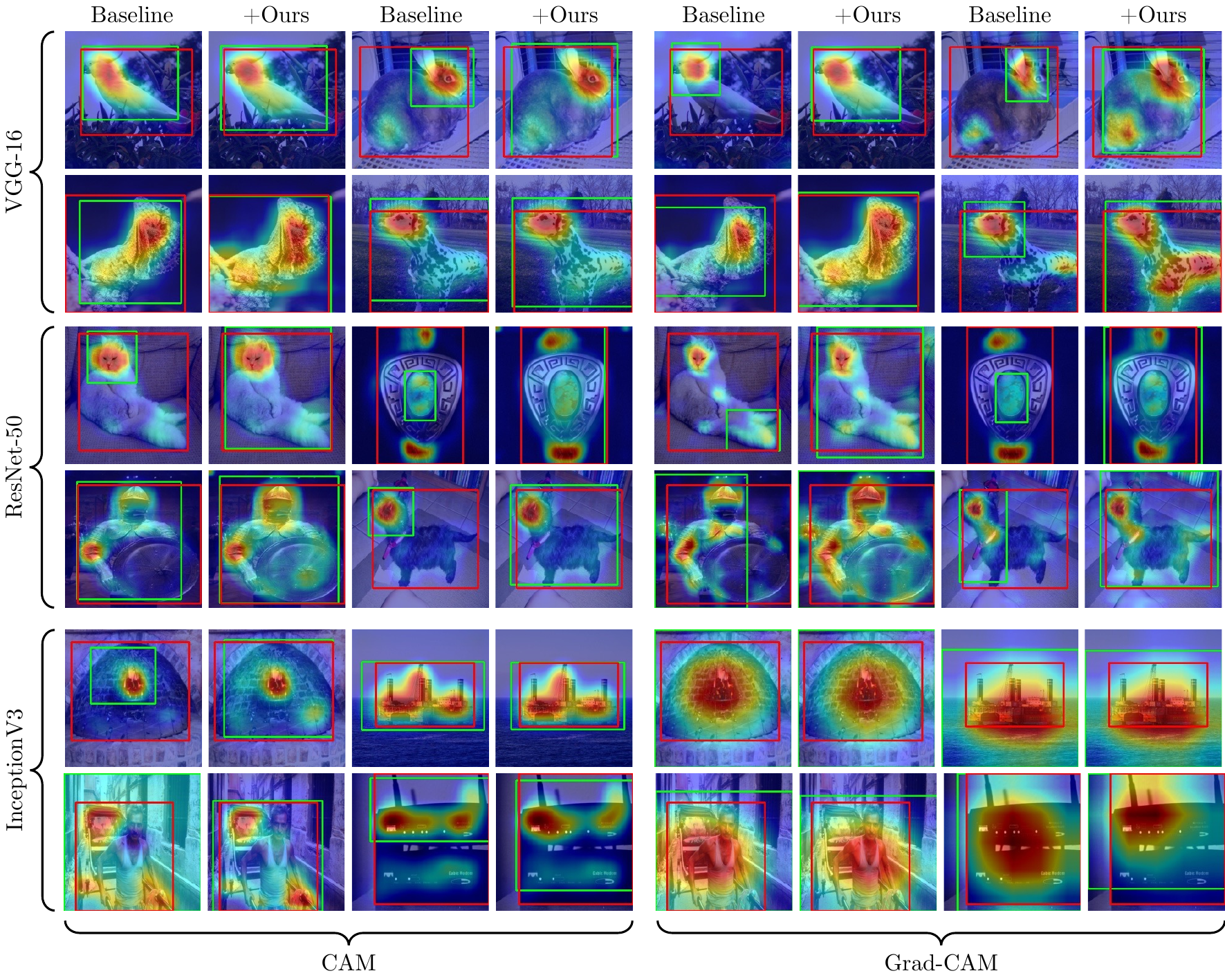}
\end{center}
\caption{
Additional qualitative WSOL examples on ImageNet-1K using VGG-16 (top), ResNet-50 (middle), and InceptionV3 (bottom). Predicted bounding boxes are shown in green, and ground-truth boxes in red.
}
\label{fig:qual_supp_imagenet}
\end{figure*}

\begin{figure*}[t!]
\begin{center}
\includegraphics[width=\textwidth]{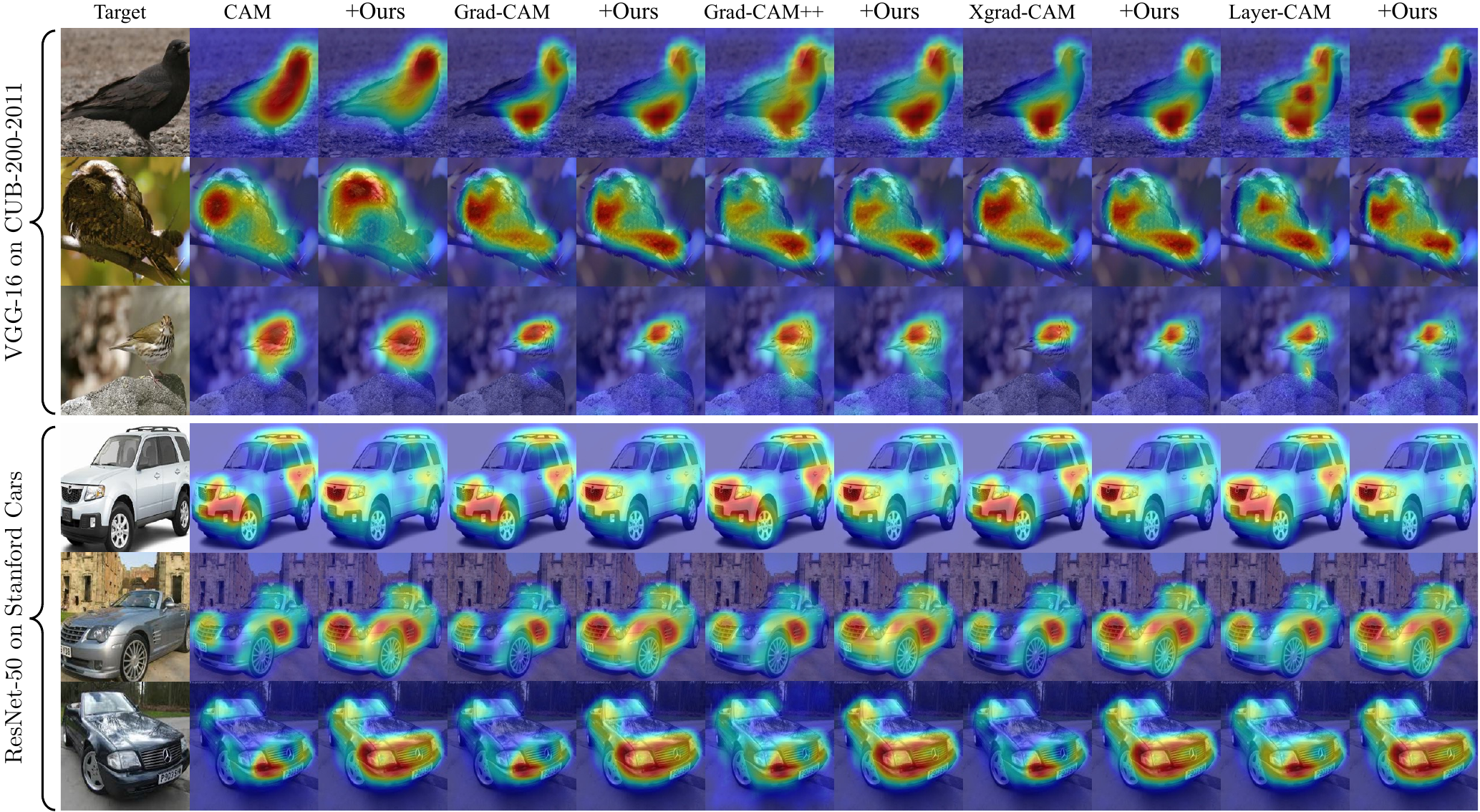}
\end{center}
\caption{
Additional qualitative explanation examples on fine-grained datasets: VGG-16 on CUB-200-2011 (top) and ResNet-50 on Stanford Cars (bottom).
}
\label{fig:qual_supp_xai}
\end{figure*}
\end{document}